\pdfoutput=1
\documentclass[10pt,twocolumn,letterpaper]{article}

\usepackage{preprint}
\usepackage{amsthm}
\usepackage{mathtools}
\usepackage{bm}
\usepackage{multirow}
\usepackage{cuted}
\usepackage{currfile}
\usepackage{caption}
\usepackage{listings}
\usepackage{colortbl}

\newcommand{\PH}[1]{\textcolor{gray}{\textbf{[}#1\textbf{]}}}

\newcommand{\Tri}{\mathcal{T}}
\newcommand{\Eland}{\mathcal{E}_{\mathrm{land}}}

\newcommand{\sg}{\operatorname{sg}}
\DeclareMathOperator{\Cov}{Cov}

\newcommand{\E}{\mathbb{E}}

\newcommand{\vc}{v_{\mathrm{c}}}
\newcommand{\xh}{\hat{x}}

\usepackage{xcolor}
\usepackage[most]{tcolorbox}
\definecolor{uwmblue}{HTML}{1F5BA8}
\definecolor{uwmgreen}{HTML}{23795A}
\definecolor{uwmgray}{HTML}{646A6E}
\definecolor{uwmsand}{HTML}{A47B36}

\newtcbox{\StTag}[1]{on line, nobeforeafter, tcbox raise base, arc=1.6pt, boxrule=0pt, boxsep=0pt,
  left=1.9pt, right=1.9pt, top=0.9pt, bottom=0.5pt, colback=#1, opacityback=0.13,
  colframe=#1, opacityframe=0, fontupper=\sffamily\bfseries\footnotesize\color{#1!85!black}}
\newcommand{\StHead}[3]{\StTag{#1}{#2}\ifstrempty{#3}{\hspace{0.45em}}{\hspace{0.4em}\textbf{#3.}\hspace{0.45em}}}
\newlength{\StSkip}
\newenvironment{StStatement}[5]{%
  \par\addvspace{\StSkip}\noindent\refstepcounter{#1}%
  \StHead{#3}{#2~\csname the#1\endcsname}{#4}#5\ignorespaces}
  {\par\addvspace{\StSkip}}

\newenvironment{theorembox}[2][]{\begin{StStatement}{sptheorem}{Theorem}{uwmblue}{#2}{\itshape}}{\end{StStatement}}
\newenvironment{propositionbox}[2][]{\begin{StStatement}{spproposition}{Proposition}{uwmgreen}{#2}{\itshape}}{\end{StStatement}}
\newenvironment{definitionbox}[2][]{\begin{StStatement}{spdefinition}{Definition}{uwmsand}{#2}{\upshape}}{\end{StStatement}}
\newcounter{assumption}

\newenvironment{corollarybox}[2][]{\begin{StStatement}{spcorollary}{Corollary}{uwmblue}{#2}{\itshape}}{\end{StStatement}}
\newenvironment{lemmabox}[2][]{\begin{StStatement}{splemma}{Lemma}{uwmgreen}{#2}{\itshape}}{\end{StStatement}}
\theoremstyle{definition}

\newcommand{\SecRefFormat}[1]{%
  \crefformat{#1}{\S##2##1##3}\Crefformat{#1}{\S##2##1##3}%
  \crefrangeformat{#1}{\S\S##3##1##4--##5##2##6}\Crefrangeformat{#1}{\S\S##3##1##4--##5##2##6}%
  \crefmultiformat{#1}{\S\S##2##1##3}{ and~##2##1##3}{, ##2##1##3}{ and~##2##1##3}%
  \Crefmultiformat{#1}{\S\S##2##1##3}{ and~##2##1##3}{, ##2##1##3}{ and~##2##1##3}%
  \crefrangemultiformat{#1}{\S\S##3##1##4--##5##2##6}{ and~##3##1##4--##5##2##6}{, ##3##1##4--##5##2##6}{ and~##3##1##4--##5##2##6}%
  \Crefrangemultiformat{#1}{\S\S##3##1##4--##5##2##6}{ and~##3##1##4--##5##2##6}{, ##3##1##4--##5##2##6}{ and~##3##1##4--##5##2##6}}
\AtEndPreamble{%
  \SecRefFormat{section}\SecRefFormat{subsection}\SecRefFormat{subsubsection}%
  \crefname{sptheorem}{Thm.}{Thms.}\Crefname{sptheorem}{Thm.}{Thms.}%
  \crefname{spproposition}{Prop.}{Props.}\Crefname{spproposition}{Prop.}{Props.}%
  \crefname{spcorollary}{Cor.}{Cors.}\Crefname{spcorollary}{Cor.}{Cors.}%
  \crefname{splemma}{Lem.}{Lems.}\Crefname{splemma}{Lem.}{Lems.}%
  \crefname{spdefinition}{Def.}{Defs.}\Crefname{spdefinition}{Def.}{Defs.}%
  \crefname{assumption}{Assump.}{Assumps.}\Crefname{assumption}{Assump.}{Assumps.}%
  \crefname{remark}{Rem.}{Rems.}\Crefname{remark}{Rem.}{Rems.}%
}

\definecolor{isth}{HTML}{2563EB}
\definecolor{isl}{HTML}{BA0C2F}
\definecolor{nomc}{HTML}{8C6D3A}

{}}

{\PH{--}}}

{}}
\graphicspath{{figs/}}

\definecolor{codebg}{HTML}{F3F6FB}
\definecolor{codecm}{HTML}{6B7280}
\newtcblisting{codebox}{enhanced, listing only, frame hidden, boxrule=0pt, arc=2pt, colback=codebg,
  borderline west={1.6pt}{0pt}{isth}, left=7pt, right=3pt, top=1.5pt, bottom=1.5pt, boxsep=0pt,
  before skip=5pt plus 1pt, after skip=6.5pt plus 1pt,
  listing options={language=Python, basicstyle=\ttfamily\scriptsize, columns=fullflexible, keepspaces=true,
    showstringspaces=false, commentstyle=\color{codecm}\itshape, keywordstyle=\color{black},
    emph={isth}, emphstyle=\color{isth}\bfseries, literate={~}{{$\sim$}}1, aboveskip=0pt, belowskip=0pt}}

\newcommand{\brno}{\textcolor{black!60}{$\times$}}
\newcommand{\brown}{\checkmark}
\newcommand{\bryes}{\textcolor{isth}{\checkmark}}

\newcommand{\rf}[1]{\textcolor{black!60}{#1}}

\usepackage{tabularx}
\definecolor{ovours}{HTML}{E8F0FE}
\definecolor{ovoursk}{HTML}{CBDCFB}
\definecolor{ovkey}{HTML}{F1F1F1}
\newcommand{\tst}{\rule{0pt}{2.55ex}}
\newcommand{\bst}{\rule[-1.05ex]{0pt}{0pt}}

\makeatletter
\renewcommand\paragraph{\@startsection{paragraph}{4}{\z@}%
  {1ex \@plus .5ex \@minus .2ex}{-1em}{\normalfont\normalsize\bfseries}}
\makeatother

\definecolor{linkblue}{rgb}{0.21,0.49,0.74}
\usepackage[pagebackref,breaklinks,colorlinks,allcolors=linkblue]{hyperref}

\input{supp_labels}

\title{\raisebox{-.35ex}{\includegraphics[height=2.15ex]{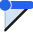}} Isthmus: Bridging Supervision Islands\\ Restores Uncertainty and Stability in One-Step Flows}

\author{Luming Yang\textsuperscript{1,*}, Siqing Li\textsuperscript{1,*}, Haoxian Liu\textsuperscript{*}, Zhenyu Bu\textsuperscript{1}, Jiazheng Zhu\textsuperscript{1}, Alper Yilmaz\textsuperscript{1}\\
\textsuperscript{1}The Ohio State University}

\hypersetup{pdftitle={Isthmus: Bridging Supervision Islands Restores Uncertainty and Stability in One-Step Flows},pdfauthor={Luming Yang, Siqing Li, Haoxian Liu, Zhenyu Bu, Jiazheng Zhu, Alper Yilmaz},pdfsubject={},pdfkeywords={},pdfcreator={},pdfproducer={}}
\pdftrailerid{}

\begin{document}

\twocolumn[{%
\renewcommand\twocolumn[1][]{#1}%
\maketitle
\begin{center}
\includegraphics[width=\linewidth]{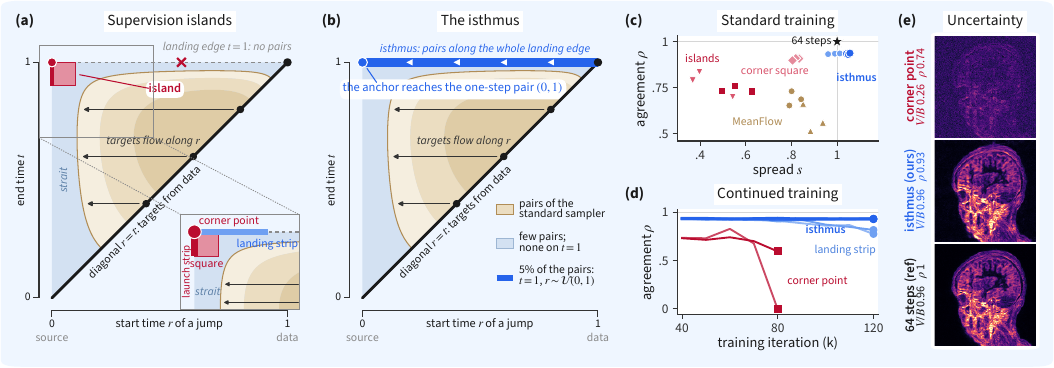}
\captionof{figure}{\textbf{Supervision islands and the isthmus.}
\textbf{(a)}~Each point is a jump from start time $r$ to end time $t$; one-step sampling is the top-left corner, from source to data. The standard time sampler~\cite{geng2025meanflow,geng2025imf} places no pairs on $t=1$, so supervision added only at the corner has no path to the diagonal. We call it a \emph{supervision island}.
\textbf{(b)}~An \emph{isthmus} spends $5\%$ of the pairs on the landing edge, linking the corner to the diagonal.
\textbf{(c)}~Training run on brain MRI comparing one-step with 64-step samples.
\textbf{(d)}~The corner point breaks down at $80$k iterations.
\textbf{(e)}~Per-pixel standard deviation (\textit{s.d.}) of $16$ samples. The corner point loses the uncertainty of 64-step reference, and the isthmus keeps it.}
\label{fig:teaser}
\end{center}
}]

\begin{NoHyper}
\renewcommand\thefootnote{}%
\footnotetext{\raggedright\textsuperscript{*}Equal contribution. Correspondence: Luming Yang \texttt{yang.7670@osu.edu}, Alper Yilmaz \texttt{yilmaz.15@osu.edu}. Photogrammetric Computer Vision Laboratory (PCVLab).}%
\end{NoHyper}

\begin{abstract}
One-step flow maps are increasingly used as posterior samplers, where the point of sampling is the uncertainty that the samples carry. We find that the one-step sampler loses this uncertainty and becomes less stable. Compared with multi-step samples of the same network, its samples lose variance and decorrelate. Supervising the one-step jump directly makes this worse; the samples collapse further toward the conditional mean and break down when training is extended. We trace both failures to the locations where training places its pairs of start and end times. Only pairs on the diagonal of the time triangle receive targets from data, and the pair used for one-step sampling is either left unsupervised or supervised on an \emph{island}, a set of pairs that the self-consistency identity does not connect to the diagonal, where the objective neither determines the one-step map nor anchors its training.
We bridge the island with an \emph{isthmus}, which places $5\%$ of the training pairs along the whole edge of the triangle that joins the one-step pair to the diagonal. The error of the one-step map is then bounded by the self-consistency residual along this edge. 
With the network, the loss and the inference code unchanged and at no added cost, the isthmus improves one-step sampling across several flow-map objectives, in MRI reconstruction, face inpainting, and unconditional generation on CIFAR-10 and ImageNet $32\times32$.
On brain MRI, it restores one-step calibration to $0.96$ and the correlation with multi-step samples to $0.93$; both vary far less across seeds, and the agreement holds when training is extended to three times its length.
\end{abstract}

\section{Introduction}
\label{sec:intro}

Diffusion and flow matching~\cite{ho2020ddpm,song2021sde,lipman2023flow,liu2023rectified,albergo2023stochastic} generate a sample over many small steps. 
Flow-map models~\cite{song2023consistency,song2024ict,lu2025scm,frans2025shortcut,boffi2025fmm,boffi2025buildcm,sabour2025ayf} instead learn the jump itself, so sampling can take as few as one network evaluation.
Mean flow~\cite{geng2025meanflow} trains such a model with a self-consistency identity and reaches strong one-step quality without distillation. Mean-flow training has since become a default choice for few-step generation~\cite{geng2025imf,zhang2026alphaflow,lee2026dmf,lu2026pmf} and has been adopted in robot control~\cite{sheng2026mp1} and inverse problems~\cite{shoushtari2026snap,shi2026nullflow,mammadov2026vfm}. In these, diffusion models have made posterior sampling a standard practice~\cite{chi2023diffusion,chung2023dps,jalal2021robust,chung2022scoremri}, the spread of the samples is read as uncertainty and diversity~\cite{edupuganti2021uq,luo2023bayesian}, and a single evaluation is expected to return a \emph{sample from the posterior}.

Whether a fast sampler preserves this spread is a known concern. Conditional GANs collapse it unless regularized~\cite{adler2018deep,ohayon2021high,bendel2023rcgan}, one-step generators miss rare modes~\cite{lin2026subflow}, and one-step policies need a dispersive term to keep their outputs apart~\cite{sheng2026mp1,wang2025disperse}. For flow models, the spread has been analyzed only for multi-step sampling~\cite{sharma2026epistemic,xing2026single,han2026uqflow}, which leaves open whether a one-step jump of the same model keeps it.
This question can be answered without ground truth: a flow-map network can jump from the source to the data in one step and can also integrate its flow in many short steps, so its one-step and multi-step samplers should agree, both in their samples and in their spread.

In our experiments, they do not agree with each other (\cref{fig:teaser}c--e). First, the one-step sampler loses the uncertainty of the model. The 64-step sampler is calibrated in the sense of~\cite{bendel2023rcgan}, while the one-step samples of the standard recipe are already under-dispersed (calibration $0.67$, where $1$ is exact). A direct idea is to supervise the one-step jump~\cite{tu2026temf, shoushtari2026snap}, but this makes it worse and collapses the samples toward the conditional mean (calibration $0.26$). Second, the one-step sampler loses agreement with the multi-step sampler and becomes unstable. In the standard recipe it is decorrelated from the 64-step sampler (correlation $0.58$), and when training is extended to two to four times its length, the collapsed one-step samplers break down while the 64-step sampler of the same network stays intact (\cref{fig:teaser}d).

Existing work on mean flows does not explain this gap. The instability of their training is well documented, but the remedies mainly act on the loss~\cite{geng2025imf,lu2026variance,zhang2026stablemf,li2026emf}. 
\citet{tu2026temf} report that results degrade when over-sampling the one-step jump, suggesting pair placement matters.
On the theory side, flow map is proven to be the unique minimizer when the loss covers the whole triangle of time pairs~\cite{boffi2025fmm,boffi2025buildcm}, however, practical samplers cover parts of it sparsely. To the best of our knowledge, no prior work has examined what the objective determines for these rarely sampled time pairs.

We show that both symptoms come from where the training pairs are placed. Training draws pairs of a start time $r$ and an end time $t$ from this triangle (\cref{fig:teaser}a). Only the diagonal $r=t$ receives targets from the data, and the mean-flow identity passes this information along rows of constant end time. One-step sampling jumps directly from the source at $r=0$ to the data at $t=1$. Its time pair $(0,1)$ lies at one end of the \emph{landing edge}, the row $t=1$ in \cref{fig:teaser}a, on which the standard sampler~\cite{geng2025meanflow,geng2025imf} places no training pairs.

Supervision added at the corner alone is therefore a \emph{supervision island}, cut off from things that the diagonal reaches. 
This is much like common noise schedules in diffusion, which never train the state that sampling starts from~\cite{lin2024common}. 
An island has two failures. First, its solution is not unique. The true flow followed by an arbitrary map satisfies the identity exactly (\cref{thm:island}), including a contraction toward the conditional mean, so the objective cannot distinguish a sampler from a regressor. 
Second, its training is not anchored. The targets at the corner are built from outputs at pairs that no training pair covers. For a linear model, training on only part of the landing edge can diverge (\cref{sec:faces}), as in off-policy temporal-difference (TD) learning~\cite{wang2024sct,baird1995residual,tsitsiklis1997analysis}.

The non-uniqueness and the lack of anchoring call for the same remedy. \emph{The error of the one-step map is bounded by the residual of the identity along the landing edge, so training controls this error whenever its rows cover the whole edge} (\cref{thm:stability}). An \emph{isthmus} is a thin set of training pairs that covers this edge (\cref{fig:teaser}b); in its simplest form, $5\%$ of the pairs are drawn with $t=1$ and $r\sim\mathcal{U}(0,1)$. 
The change is confined to the time sampler and leaves the network, the loss, and inference unchanged at no added cost.

Our contributions are as follows.
\begin{itemize}[$\triangleright$,leftmargin=*,itemsep=1pt,topsep=2pt]
\item \textbf{Cause.} We show that the self-consistency objective determines a jump only as far as training connects it to the diagonal. Under standard time samplers the one-step jump has no such connection, so it is fixed only up to an arbitrary map (\cref{thm:island}) and its targets rest on untrained pairs, which explains both symptoms and why supervising the corner alone makes them worse (\cref{sec:theory}).
\item \textbf{Criterion and method.} Extending the Eulerian bound of flow-map matching~\cite{boffi2025fmm}, we bound the one-step error by the residual of the identity along the landing edge (\cref{thm:stability}), so the error is controlled once training covers this edge. The isthmus is the simplest sampler that does so with only one-line change (\cref{sec:method}).
\item \textbf{Evidence.} Experiments that change only the time sampler confirm the analysis. On unconditional CIFAR-10~\cite{krizhevsky2009cifar} and ImageNet $32\times32$~\cite{chrabaszcz2017imagenet32} with the official MeanFlow code, the isthmus restores the agreement and the spread of the samples, and training is more stable. The same repair carries over to posterior sampling for brain and knee MRI~\cite{zbontar2018fastmri} and face inpainting~\cite{liu2015celeba} (\cref{sec:experiments}).
\end{itemize}

\section{Related work}
\label{sec:related}

\paragraph{One-step flow maps.}
Consistency models~\cite{song2023consistency,lu2025scm} train jumps that end at the data, while consistency trajectory~\cite{kim2024ctm}, shortcut~\cite{frans2025shortcut} and inductive moment matching~\cite{zhou2025imm} models train jumps between arbitrary times. Flow-map matching~\cite{boffi2025fmm,boffi2025buildcm} unifies these objectives and proves that the flow map is the unique minimizer when the loss covers the whole triangle of times. 
Mean flows~\cite{geng2025meanflow,geng2025imf} are its Eulerian member, trained with a continuous $(r,t)$ sampler. Some recent mean-flow recipes place pairs on or near the landing edge as a design choice~\cite{lee2026dmf,wang2026tim,zhang2025remeanflow}. Our \cref{thm:bridge} rewrites the proof of the Eulerian bound~\cite{boffi2025fmm} as an identity weighted by the time sampler (\cref{S-sec:proofs}), showing when such pairs suffice.

\paragraph{Stability and spread of one-step samplers.}
Recent work stabilizes mean-flow training by changing the loss~\cite{geng2025imf,lu2026variance,tu2026temf,zhang2026stablemf,li2026emf}. One-step flow maps are used for posterior sampling~\cite{shoushtari2026snap}, where the spread of the samples is read as uncertainty and should be calibrated; we measure calibration with the criterion that \citet{bendel2023rcgan} proposed for conditional GANs. We trace both the instability and the loss of spread to a cause outside the loss, namely where the time sampler places its pairs. 
An extended discussion is in \cref{S-sec:related-ext}.

\section{Analysis}
\label{sec:theory}
\label{sec:prelim}

\begin{figure*}[t]
\centering
\includegraphics[width=\linewidth]{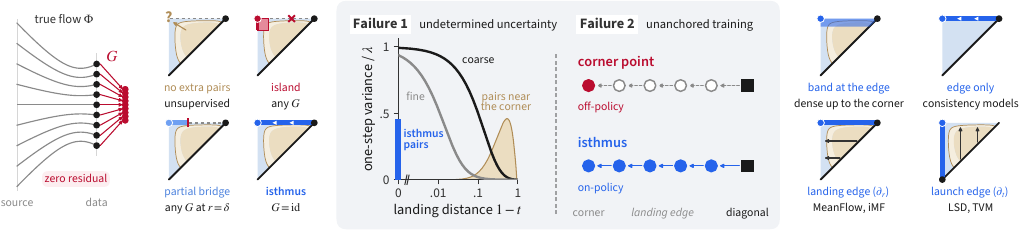}
\caption{\textbf{Theory of supervision islands.}
\textbf{Left.} Off the diagonal, the true flow $\Phi$ followed by any map $G$ has zero residual, even a $G$ that collapses every sample to the mean; only training pairs along the whole landing edge force $G=\mathrm{id}$ (\cref{thm:island}).
\textbf{Middle.} Failure~1: rows near the corner end before the data, where a one-step prediction has the right mean and too little spread (\cref{prop:layer}). Failure~2: the corner's target is built from the network's outputs along the landing edge, which corner-only training does not sample (off-policy) and the isthmus does (on-policy).
\textbf{Right.} A derivative in $r$ needs the landing edge, a derivative in $t$ the launch edge (\cref{prop:direction}).}
\label{fig:theory}
\end{figure*}

\subsection{Flow maps on the time triangle}
\label{sec:triangle}
We follow the flow-map convention~\cite{boffi2025buildcm}, in which time $0$ is the source and time $1$ is the data. MeanFlow~\cite{geng2025meanflow} runs time the other way, and its $(r,t)$ are our $(1-t,1-r)$ (\cref{S-sec:conventions}). Given a condition $y$ (omitted below), a source $x_0\sim q$ and data $x_1\sim p$ are independent, and $z_\tau=(1-\tau)x_0+\tau x_1$ has law $p_\tau$. The marginal velocity is $v(z,\tau)=\E[x_1-x_0\mid z_\tau=z]$ and $\Phi_{r\to t}$ is its flow (assumptions in \cref{S-sec:proofs}). The true average velocity $u^*$ from $r$ to $t\ge r$ is defined by
\begin{equation}
(t-r)\,u^*(z_r,r,t)=\int_r^t v(z_\tau,\tau)\,d\tau,
\label{eq:avgvel}
\end{equation}
with the integral along the trajectory through \((z_r,r)\). For a field \(u\), like a network, the jump ends at \(X_u(z,r,t)=z+(t-r)\,u(z,r,t)\), and a one-step sample is \(\xh=X_u(z,0,1)\), \(z\sim q\). By \eqref{eq:avgvel}, \(u^*\) satisfies the mean-flow identity~\cite{geng2025meanflow}
\begin{equation}
u=v+(t-r)\big(\partial_z u\,v+\partial_r u\big).
\label{eq:identity}
\end{equation}
A network $u$ is regressed on $T=\vc+(t-r)\,\sg[\partial_z u\,\vc+\partial_r u]$, with $\vc=x_1-x_0$, stop-gradient $\sg$, and one time pair $(r,t)$ per training example from a time sampler $\pi$.

Time pairs lie in the triangle $\Tri=\{0\le r\le t\le 1\}$ (\cref{fig:teaser}a).
The pairs on the \emph{diagonal} \(r=t\), where the target is \(\vc\) and does not involve the network's own output, are the flow-matching pairs.
The \emph{corner} $(0,1)$ at the top left is the pair used for one-step sampling (a jump from the source to the data).
A \emph{row} is a line of constant end time $t$ and reaches the diagonal at its right end.
The row through the corner, $t=1$, is the \emph{landing edge} $\Eland$, on which jumps end on the data. The column through it, $r=0$, is the \emph{launch edge}.

MeanFlow and its successors~\cite{geng2025meanflow, geng2025imf,zhang2026alphaflow} sample the pair as the minimum and maximum of two logit-normal times and set a fraction of the pairs on the diagonal. Off the diagonal, this sampler has a density that gives $\Eland$ zero mass and is small near the corner (\cref{fig:teaser}a).

\subsection{Residual along a row}
\label{sec:free}
Let $R=u-v-(t-r)\big(\partial_z u\,v+\partial_r u\big)$ be the residual of \eqref{eq:identity}, where $v$ represents the unobserved velocity. The training loss equals $\E\|R\|^2$ plus a variance term depending on $u$, and $R$ vanishes almost surely at a population stationary point of an expressive model (\cref{S-lem:fixedpoint}). Proofs are in \cref{S-sec:proofs}.
\begin{lemmabox}{Endpoint conservation}
\label{lem:conserve}
Along any trajectory $z_r=\Phi_{0\to r}(z_0)$,
\begin{equation}
\tfrac{d}{dr}X_u(z_r,r,t)=-R(z_r,r,t) .
\label{eq:conserve}
\end{equation}
\end{lemmabox}
The identity holds if and only if the predicted endpoint is constant along trajectories. It differentiates only in the start time, so it relates pairs on the same row and does not couple rows. On a row, the boundary condition is at the diagonal end, where $X_u=z$ for bounded $u$.

\begin{theorembox}{Off the diagonal the jump is free}
\label{thm:island}
Fix an end time $t$ and start times $[a,b]$ with $b<t$. A $C^1$ field $u$ has zero residual on $[a,b]\times\{t\}$ if and only if
\begin{equation}
X_u(\cdot,r,t)=G\circ\Phi_{r\to t}\qquad\text{for all } r\in[a,b]
\label{eq:gphi}
\end{equation}
with a map $G$ independent of $r$, and every $C^1$ map $G$ is attained. If instead $b=t$ and $u$ is bounded, then $G=\mathrm{id}$.
\end{theorembox}
A piece of a row that stops short of the diagonal fixes its jumps up to a common map $G$ at zero residual (\cref{fig:theory}, left). $G\equiv c$ gives the collapse $u=(c-z)/(t-r)$, which satisfies \eqref{eq:identity} for every velocity field, and with $c=\E[x_1\mid y]$ the one-step sampler is the regression estimator. This field is unbounded as $r\to t$ and cannot reach the diagonal.

\subsection{Error bound}
\label{sec:bridge}
\Cref{thm:island} assumes zero residual, in which case a network continuous in $t$ and trained on a density is determined everywhere (\cref{S-prop:exact}).
However, a trained network has a non-zero residual, and then every $G$ is attained at the corner at an arbitrarily small residual whenever the training pairs leave a gap on the landing edge (\cref{S-prop:ident}). The training mass between a pair and the diagonal can also be very small. Integrating \eqref{eq:conserve} along the row accounts for both.

\begin{definitionbox}{}
\label{def:island}
The \emph{bridge} of a pair $(r_0,t)$ is the part of its row between the pair and the diagonal. Let $\pi_t(r)$ be the density in $r$ of the training mass on row $t$ itself, non-zero only if the end time has an atom at $t$. The pair is \emph{bridged} if $C_\pi(r_0,t)=\int_{r_0}^{t}dr/\pi_t(r)\in(0,\infty]$ is finite. Otherwise it is \emph{unsupervised} if no pairs are added for it. Pairs added on a set containing it, without bridging it, form a \emph{supervision island}. An island on the bridge is a \emph{partial bridge} (\cref{fig:theory}).
\end{definitionbox}
Being bridged is a guarantee from the sampler alone regardless of the network. A density bridges no off-diagonal pair and not even the short jumps of multi-step sampling; continuity in $t$ controls those (\cref{S-prop:band}).

\begin{theorembox}{Bridge criterion}
\label{thm:bridge}
\label{thm:stability}
Let $r_0<t$, and let $u(\cdot,\cdot,t)$ be $C^1$ and bounded for $r\in[r_0,t)$. For every $z$, with $z_r=\Phi_{r_0\to r}(z)$,
\begin{equation}
X_u(z,r_0,t)-\Phi_{r_0\to t}(z)=\int_{r_0}^{t}R(z_r,r,t)\,dr .
\label{eq:stab-id}
\end{equation}
Hence $e^2(r_0,t)=\E_{p_{r_0}}\|X_u(\cdot,r_0,t)-\Phi_{r_0\to t}\|^2$, the mean-square error of the jump, bounds the squared $W_2$ distance between $X_u(\cdot,r_0,t)_\#p_{r_0}$ and $p_t$ and satisfies
\begin{equation}
e^2(r_0,t)\le C_\pi(r_0,t)\int_{r_0}^{t}\!\pi_t(r)\,\E_{p_r}\|R(\cdot,r,t)\|^2\,dr .
\label{eq:stab}
\end{equation}
\end{theorembox}
The error of a jump is the residual accumulated along its bridge. With zero residual on the whole bridge, the error vanishes, which recovers the case $b=t$ of \cref{thm:island}. Otherwise, the integral in \eqref{eq:stab} is the part on the bridge of $\E_\pi\|R\|^2$, the mean-square residual over the training pairs. The bound is a row-wise form of the Eulerian bound of~\cite{boffi2025fmm}, weighted by the time sampler; for $C_\pi=\infty$ it is void and does not mean that the error is large.

\begin{corollarybox}{The one-step corner}
\label{cor:sampler}
The bridge of the corner is the landing edge (\cref{fig:theory}).
(i)~Training pairs on the whole edge with $\pi_1\ge c>0$ give $C_\pi(0,1)\le1/c$, so \eqref{eq:stab} bounds the one-step error by the residual on the edge.
(ii)~Pairs only on $[0,\delta]$, $\delta<1$, form a partial bridge: $C_\pi(0,1)=\infty$, and the jumps share an arbitrary $G$ at zero residual; $\delta=0$ is the corner alone.
(iii)~Without pairs on the edge, $C_\pi(0,1)=\infty$ and the population objective does not depend on $u$ on $\Eland$. Other rows constrain the corner only through the continuity of the network in $t$.
\end{corollarybox}

In case~(iii) the corner is reached through neighboring rows. If the learned and the true jumps are $L$-Lipschitz in the end time, then $e(0,1)\le e(0,t')+2L(1-t')$ (\cref{S-cor:stability}(ii), \cref{S-prop:band}). This bound is useful only when the rows stay dense up to the edge. For the standard sampler, the row mass vanishes toward $t=1$, and only $1.2\times10^{-5}$ of its pairs lie in the band $r\le0.3$, $t\ge0.98$ (\cref{S-tab:family}). We call this nearly empty region the \emph{strait} (\cref{fig:teaser}a). Across it, the bound rests on $L$, which the objective does not control. A bridge removes this dependence. A partial bridge (case~ii) moves the arbitrary $G$ from the corner to the pair $(\delta,1)$, and only the full edge removes it.

\subsection{Possible consequences}
\label{sec:faces}
Without a bridge, the one-step map is left to the network's extrapolation, which can differ between training runs (\cref{S-prop:ident}) and affect both the value the corner takes and the stability of its training. The multi-step sampler should be less affected, since its short jumps lie next to the diagonal where the continuity bound is tight.

\paragraph{Failure 1: undetermined uncertainty.}
The supervised jumps that lie nearest to the corner start close to the source but stop at some $t<1$, short of the data.
\begin{propositionbox}{}
\label{prop:layer}
Let $F_t(z)=z+u^*(z,0,t)$, $t\in(0,1]$, be the one-step prediction read off the true jump \eqref{eq:avgvel} from the source to time $t$. For any data distribution with finite second moments,
(i)~$\E_z F_t(z)=\E[x_1]$,
(ii)~$\operatorname{tr}\Cov F_t\le\operatorname{tr}\Cov x_1$, with equality at $t=1$,
(iii)~every average of $F_t$ over $t$ satisfies (i) and (ii).
Moreover, $F_t\to\E[x_1]$ as $t\to0$.
\end{propositionbox}
\Cref{prop:layer} describes true jumps. An island trains those jumps next to the corner. If the one-step jump takes its value from them, which \eqref{eq:identity} does not enforce, the samples keep their mean and contract. A corner supervised alone has as a stationary value the regression target plus a derivative of the network that no other pair constrains (\cref{S-prop:value}), which draws it toward regression. An unsupervised corner is not tied to the multi-step sampler at all. In the linear-Gaussian model of \cref{S-sec:lg}, the contraction in a direction of variance $\lambda$ starts once the landing distance $1-t$ exceeds about $\sqrt\lambda$ (\cref{fig:theory}). Therefore, directions of small variance such as fine detail contract first. The sign of the error is not fixed since an expanding $G$ is equally admissible (\cref{thm:island}). The residual can be zero and does not reveal this error.

\paragraph{Failure 2: unanchored training.}
The corner target contains the network's \(r\)-derivative, its variation along the landing edge. 
Thus, stop-gradient training behaves like temporal-difference (TD) learning on a chain from the corner to the diagonal~\cite{wang2024sct}, visiting the edge uniformly, as illustrated in \cref{fig:theory} (\cref{S-rem:td}). 
For a linear model trained on the landing edge alone (\cref{S-prop:dyn}), uniform pairs on the whole edge sample the states of this chain (on-policy), and the expected training dynamics provably contracts to one solution from every initialization, as in linear TD learning~\cite{tsitsiklis1997analysis,bertsekas1996neuro}. An island bootstraps from states it does not sample (off-policy), so its targets are not anchored: a strip can diverge~\cite{baird1995residual}, and at the corner alone some changes of the one-step value leave the update unchanged. For the network this suggests less stability under long training.

\section{Isthmus Sampling}
\label{sec:method}

\label{sec:rule}
Both failures arise without a bridge; hence, we add one.
\begin{definitionbox}{Isthmus}
\label{def:isthmus}
A time sampler \emph{has an isthmus} if the pair used for sampling is bridged (\cref{cor:sampler}(i)). An \emph{isthmus sampler} is a mixture $\pi=(1-\pi_{\mathrm{isth}})\,\pi_{\mathrm{base}}+\pi_{\mathrm{isth}}\,\pi_{\mathrm{bridge}}$ with a share $\pi_{\mathrm{isth}}\in(0,1]$ and a distribution $\pi_{\mathrm{bridge}}$ on the bridge under which this pair is bridged.
\end{definitionbox}
Our default is the \emph{uniform isthmus}. With probability $\pi_{\mathrm{isth}}=5\%$, a training example's time pair is replaced by
\begin{equation}
r\sim\mathcal{U}(0,1),\qquad t=1,
\label{eq:isthmus}
\end{equation}
which is a jump from a uniform start that lands on the data. In the MeanFlow convention, it reads $t\sim\mathcal{U}(0,1)$, $r=0$ (\cref{S-sec:conventions}). The analysis requires only $\pi_{\mathrm{isth}}>0$; the share is a hyperparameter, and $2.5\%$ to $10\%$ perform alike (\cref{sec:exp-anatomy}).
\begin{codebox}
r, t = base_sampler(n)        # any sampler, r <= t
isth = rand(n) < 0.05         # 5% of the examples
r = where(isth, rand(n), r)   # start ~ U(0, 1)
t = where(isth, 1.0, t)       # end on the data
\end{codebox}
\noindent Only these codes change. The new pairs are trained with the unchanged mean-flow target. The network, loss weighting, and inference are untouched. Pairs are replaced rather than added, so the cost per iteration is unchanged, and the base sampler keeps a share $1-\pi_{\mathrm{isth}}$ of the pairs.

The base sampler has no pair on $\Eland$, so $\pi_1\equiv\pi_{\mathrm{isth}}$, $C_\pi(0,1)=1/\pi_{\mathrm{isth}}$, and the share cancels in \eqref{eq:stab}:
\begin{equation*}
e^2(0,1)\le\E_{r\sim\mathcal{U}(0,1)}\E_{p_r}\|R(\cdot,r,1)\|^2\le\E_\pi\|R\|^2/\pi_{\mathrm{isth}} .
\end{equation*}
The first bound is the mean-square residual over the isthmus pairs, with a constant $1$ for any share, which enters only relative to the residual over all pairs.
The uniform density minimizes $C_\pi$ at fixed mass (\cref{S-cor:stability}(i)), that is, the constant, and matches the state distribution of the chain in \cref{sec:faces}.
A small error implies closeness to \(p_1\) in \(W_2\) and agreement with the multi-step samples (\cref{S-cor:stability}(iii)), and two training runs or checkpoints then differ at the corner by at most the sum of their edge residuals (\cref{S-prop:ident}). This is what the analysis predicts. Calibration further requires an error small relative to the posterior variance, which the bound does not provide. It is tested empirically.

\begin{table*}[t]
\centering
\footnotesize
\setlength{\tabcolsep}{2.7pt}\renewcommand{\arraystretch}{1.06}
\setlength{\aboverulesep}{0pt}\setlength{\belowrulesep}{0pt}%
\begin{tabularx}{\textwidth}{@{}l@{\hspace{9pt}}lc>{\centering\arraybackslash}X>{\columncolor{ovkey}\centering\arraybackslash}X>{\centering\arraybackslash}X>{\centering\arraybackslash}X>{\centering\arraybackslash}X>{\columncolor{ovkey}\centering\arraybackslash}X>{\centering\arraybackslash}X>{\centering\arraybackslash}X}
\toprule
 & & & \multicolumn{6}{c}{one step (NFE\,1)\tst\bst} & \multicolumn{2}{c}{\rf{same network, 64 steps}} \\
\cmidrule(lr){4-9}\cmidrule(l){10-11}
Family\tst\bst & Method & Bridge & PSNR$_\mu\!\uparrow$ & $\bm{V/B}\!\to\!1$ & CRPS$\downarrow$ & LPIPS$\downarrow$ & \makebox[0pt][c]{$V_1/V_{64}\!\to\!1$} & $\bm{\rho}\!\to\!1$ & \rf{PSNR$_\mu\!\uparrow$} & \rf{$V/B\!\to\!1$} \\
\midrule
\rf{Reference}\tst & \rf{Regression U-Net} & \rf{--} & \rf{24.74} & \rf{0.00} & \rf{3.22} & \rf{0.070} & \rf{--} & \rf{--} & \rf{--} & \rf{--} \\
\rf{(64 steps)}\bst & \rf{MeanFlow + isthmus} & \rf{--} & \rf{25.58{\tiny$\pm$.007}} & \rf{0.96{\tiny$\pm$.006}} & \rf{2.10{\tiny$\pm$.003}} & \rf{0.049{\tiny$\pm$.0002}} & \rf{1} & \rf{1} & \rf{--} & \rf{--} \\
\midrule
Eulerian\tst & MeanFlow~\cite{geng2025meanflow} & \brno & 25.00{\tiny$\pm$.004} & 0.67{\tiny$\pm$.068} & 2.56{\tiny$\pm$.049} & 0.072{\tiny$\pm$.0044} & 0.79{\tiny$\pm$.078} & 0.58{\tiny$\pm$.076} & \rf{25.56{\tiny$\pm$.028}} & \rf{0.94{\tiny$\pm$.008}} \\
(deriv.\ in $r$) & MeanFlow ($p{=}0.5$) & \brno & 25.08{\tiny$\pm$.032} & 0.56{\tiny$\pm$.032} & 2.47{\tiny$\pm$.028} & 0.064{\tiny$\pm$.0015} & 0.66{\tiny$\pm$.046} & 0.69{\tiny$\pm$.038} & \rf{25.57{\tiny$\pm$.028}} & \rf{0.93{\tiny$\pm$.006}} \\
\rowcolor{ovours}\cellcolor{white} & \quad\textbf{+ isthmus (ours)} & \bryes & 25.11{\tiny$\pm$.020} & \cellcolor{ovoursk}\textbf{0.96}{\tiny$\pm$.013} & 2.29{\tiny$\pm$.010} & 0.054{\tiny$\pm$.0003} & 1.10{\tiny$\pm$.015} & \cellcolor{ovoursk}0.93{\tiny$\pm$.004} & \rf{25.58{\tiny$\pm$.007}} & \rf{0.96{\tiny$\pm$.006}} \\
 & iMF~\cite{geng2025imf} & \brno & 24.98{\tiny$\pm$.022} & 0.69{\tiny$\pm$.064} & 2.59{\tiny$\pm$.053} & 0.075{\tiny$\pm$.0054} & 0.83{\tiny$\pm$.093} & 0.53{\tiny$\pm$.086} & \rf{25.56{\tiny$\pm$.023}} & \rf{0.92{\tiny$\pm$.007}} \\
\rowcolor{ovours}\cellcolor{white} & \quad\textbf{+ isthmus (ours)} & \bryes & 25.08{\tiny$\pm$.003} & \cellcolor{ovoursk}0.89{\tiny$\pm$.044} & 2.31{\tiny$\pm$.002} & 0.054{\tiny$\pm$.0003} & 1.03{\tiny$\pm$.053} & \cellcolor{ovoursk}0.93{\tiny$\pm$.005} & \rf{25.57{\tiny$\pm$.004}} & \rf{0.95{\tiny$\pm$.023}} \\
 & SNaP~\cite{shoushtari2026snap} & \brown & 25.22{\tiny$\pm$.025} & 0.71{\tiny$\pm$.009} & 2.28{\tiny$\pm$.009} & 0.053{\tiny$\pm$.0006} & 0.78{\tiny$\pm$.020} & 0.93{\tiny$\pm$.008} & \rf{25.56{\tiny$\pm$.025}} & \rf{0.98{\tiny$\pm$.011}} \\
\rowcolor{ovours}\cellcolor{white}\bst & \quad\textbf{+ isthmus (ours)} & \bryes & 25.16{\tiny$\pm$.006} & \cellcolor{ovoursk}\underline{0.95}{\tiny$\pm$.042} & 2.28{\tiny$\pm$.003} & 0.054{\tiny$\pm$.0007} & 1.09{\tiny$\pm$.033} & \cellcolor{ovoursk}\underline{0.94}{\tiny$\pm$.004} & \rf{25.57{\tiny$\pm$.025}} & \rf{0.96{\tiny$\pm$.021}} \\
\midrule
Lagrangian\tst & LSD~\cite{boffi2025buildcm} & \brno & 24.99{\tiny$\pm$.060} & 0.84{\tiny$\pm$.088} & 2.55{\tiny$\pm$.042} & 0.074{\tiny$\pm$.0056} & 0.93{\tiny$\pm$.116} & 0.70{\tiny$\pm$.043} & \rf{25.57{\tiny$\pm$.028}} & \rf{1.01{\tiny$\pm$.008}} \\
\rowcolor{ovours}\cellcolor{white}(deriv.\ in $t$)\bst & \quad\textbf{+ isthmus (ours)} & \bryes & 25.13{\tiny$\pm$.018} & \cellcolor{ovoursk}0.85{\tiny$\pm$.025} & 2.28{\tiny$\pm$.002} & 0.052{\tiny$\pm$.0004} & 0.90{\tiny$\pm$.027} & \cellcolor{ovoursk}\textbf{0.96}{\tiny$\pm$.001} & \rf{25.53{\tiny$\pm$.010}} & \rf{1.03{\tiny$\pm$.002}} \\
\midrule
\rf{Bridged models}\tst & \rf{Shortcut~\cite{frans2025shortcut}} & \rf{\brown} & \rf{25.08{\tiny$\pm$.018}} & \rf{0.81{\tiny$\pm$.004}} & \rf{2.30{\tiny$\pm$.005}} & \rf{0.052{\tiny$\pm$.0003}} & \rf{0.96{\tiny$\pm$.007}} & \rf{0.96{\tiny$\pm$.001}} & \rf{25.56{\tiny$\pm$.024}} & \rf{0.93{\tiny$\pm$.008}} \\
\rf{(no islands)}\bst & \rf{sCM~\cite{lu2025scm}} & \rf{\brown} & \rf{24.80{\tiny$\pm$.062}} & \rf{1.15{\tiny$\pm$.038}} & \rf{2.51{\tiny$\pm$.044}} & \rf{0.068{\tiny$\pm$.0033}} & \rf{--} & \rf{--} & \rf{--} & \rf{--} \\
\bottomrule
\end{tabularx}

\caption{\textbf{One-step posterior sampling on IXI brain MRI ($8\times$).} Bridge column distinguishes methods with no dedicated corner supervision (\brno), those that supervise the corner by design (\brown), and those that connect it to the diagonal through an isthmus (\bryes). For Shortcut the mark denotes step doubling, which trains the one-step pair. Gray columns sample the same network with 64 steps; sCM has none because its multi-step sampler re-injects noise and does not integrate the same flow. Best in \textbf{bold}, second \underline{underlined}, likewise throughout.}
\label{tab:overall}
\end{table*}

\paragraph{Other objectives and samplers.}
\label{sec:direction}
The corner's bridge is $\Eland$ because \eqref{eq:identity} differentiates in $r$. We now ask how this changes for other objectives and whether published samplers already have an isthmus.
\begin{propositionbox}{Direction}
\label{prop:direction}
For an objective that differentiates in the end time, $\partial_t X(z,r,t)=v(X(z,r,t),t)$ (Lagrangian self-distillation), rows and columns exchange roles. On a segment $\{r\}\times[a,b]$ with $a>r$, the maps with zero residual are $X(\cdot,r,t)=\Phi_{a\to t}\circ H$ with a $C^1$ map $H$. If instead $a=r$ and $u$ is bounded, then $X(\cdot,r,t)=\Phi_{r\to t}$. The bridge of the corner is then the launch edge, and training pairs on the landing edge form an island off this bridge.
\end{propositionbox}
As shown in \cref{fig:theory}, the bridge runs along the time variable that the objective differentiates (a row for \eqref{eq:identity}). For an end-time objective, the isthmus lies on the launch edge, $r=0$, $t\sim\mathcal{U}(0,1)$ in our convention. Since \eqref{eq:identity} contains no $\partial_t u$, launch-edge pairs do not bridge the MeanFlow corner. Like pairs on other rows, they act on it only through the network's continuity in $t$. On the landing edge, \eqref{eq:identity} is the continuous-time consistency condition~\cite{song2023consistency,lu2025scm}, so a consistency model places its pairs on the bridge ($\pi_{\mathrm{isth}}=1$). A bridge bounds the error of the one-step map but does not calibrate the samples by itself. Discrete targets such as step doubling of shortcut models~\cite{frans2025shortcut} are outside this analysis.

\paragraph{Existing samplers.}
\label{sec:audit}
For \eqref{eq:identity}, whether a method has an isthmus depends solely on its time sampler and can be checked without training; \cref{S-sec:family} relates the published samplers to the cases of \cref{cor:sampler} and finds no isthmus among those that draw both times from logit-normal densities (\cref{S-tab:family}).

\section{Experiments}
\label{sec:experiments}

\begin{figure*}[t]
\centering
\includegraphics[width=\linewidth]{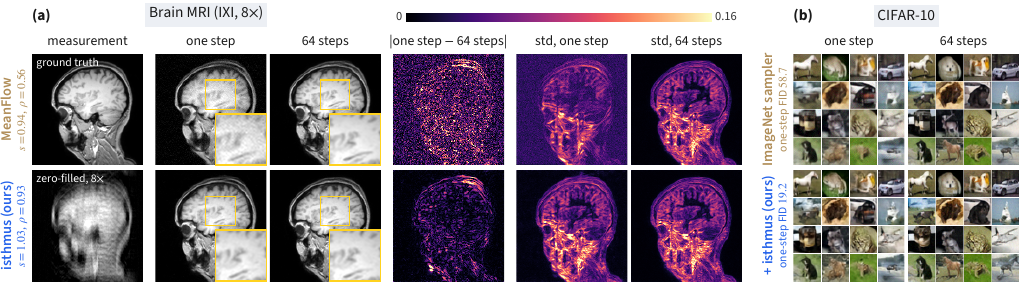}
\caption{\textbf{One-step samples with and without the isthmus.} With the isthmus, the one-step samples and their per-pixel \textit{s.d.} match those of 64 steps.
\textbf{(a)}~IXI brain MRI. Both models receive the same measurement and the same source draws, and the \textit{s.d.} is taken over $16$ samples.
\textbf{(b)}~MeanFlow with and without isthmus on unconditional CIFAR-10. All panel uses the same $16$ noises.}
\label{fig:qual}
\end{figure*}

\begin{figure*}[t]
\centering
\includegraphics[width=\linewidth]{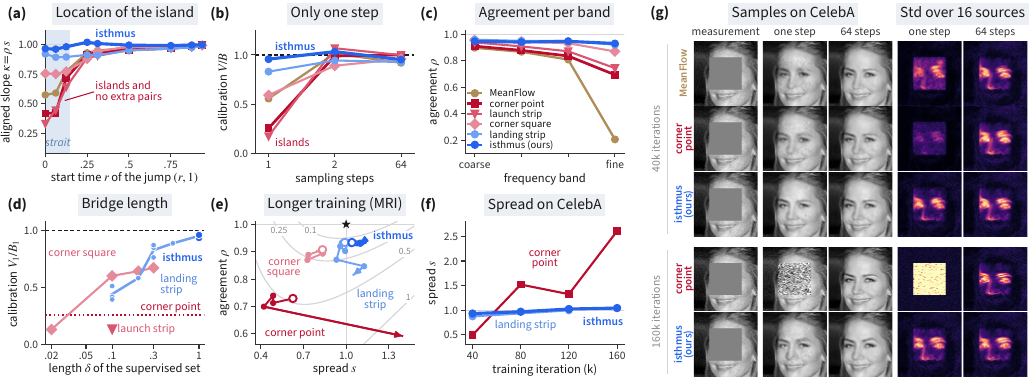}
\caption{\textbf{Analysis of islands and the isthmus.} Islands fail only for jumps that start near the source, and the corner point breaks down under longer training, while the isthmus stays consistent with 64 steps throughout. Panels (a)--(e) are on IXI brain MRI ($8\times$), and (f), (g) on CelebA.
\textbf{(a)}~Aligned slope of a single jump from start time $r$ to the data.
\textbf{(b)}~Calibration $V/B$ with $1$, $2$ and $64$ steps.
\textbf{(c)}~Paired correlation $\rho$ per frequency band.
\textbf{(d)}~Calibration $V/B$ against the length of the supervised set.
\textbf{(e)}~Paired correlation $\rho$ against spread $s$ from $40$k to $120$k iterations. Open circles mark the $40$k checkpoints, and arcs mark equal paired error $s^2-2\rho s+1$ between one-step and 64-step samples.
\textbf{(f)}~Spread $s$ from $40$k to $160$k iterations.
\textbf{(g)}~Samples for one measurement at $40$k and $160$k iterations.}
\label{fig:mech}
\end{figure*}

\label{sec:exp-setup}
\paragraph{Data and training.}
We evaluate mainly on posterior sampling for $8\times$ accelerated MRI on IXI brain slices~\cite{ixi}. A conditional U-Net is trained with the MeanFlow loss and time sampler~\cite{geng2025meanflow} for $40$k iterations, and results are means $\pm$ s.d.\ over three training seeds. 
We repeat the main comparisons on fastMRI knee data~\cite{zbontar2018fastmri} and CelebA inpainting~\cite{liu2015celeba}, and with the official MeanFlow code on unconditional CIFAR-10~\cite{krizhevsky2009cifar} and ImageNet $32\times32$~\cite{chrabaszcz2017imagenet32}. \Cref{S-sec:impl} gives the details and the training seeds behind every table.

\paragraph{Metrics.}
\label{sec:diagnostics}
We use two groups of metrics. The first group compares each one-step sampler with the 64-step sampler of the same network, which serves as its reference. The \emph{spread} $s$ is the ratio of their standard deviations. The \emph{paired correlation} $\rho$ is the correlation between a one-step sample and the 64-step sample from the same noise. The \emph{aligned slope} $\kappa=\rho s$ is the slope of a regression of the one-step sample on the 64-step sample. A one-step sampler is consistent with 64 steps when $s$, $\rho$ and $\kappa$ are close to $1$. Collapse toward the mean lowers $s$ and $\kappa$, and decorrelation lowers $\rho$. We also compute $\rho$ and $\kappa$ in frequency bands, where $\kappa_{\mathrm{c}}$ and $\kappa_{\mathrm{f}}$ denote the coarsest and the finest band. The second group compares the samples with the clean image. The \emph{calibration} $V/B$ is the variance of the samples divided by the squared error of their mean. It is $1$ for an exact posterior sampler and $0$ for a regressor~\cite{blau2018perception,bendel2023rcgan}. Subscripts give the number of sampling steps, as in $V_1/B_1$. We also report the PSNR of the sample mean (PSNR$_\mu$), the CRPS~\cite{gneiting2007strictly} ($\times10^{2}$) and LPIPS~\cite{zhang2018lpips}. \Cref{S-sec:eval} gives the estimators and the bands.

\paragraph{Baselines.}
All baselines share our U-Net, data, training budget and test set, and sample in one step. We organize the baselines into four groups according to their differentiation structure. The Eulerian family differentiates in $r$ and contains MeanFlow~\cite{geng2025meanflow}, iMF~\cite{geng2025imf} and SNaP~\cite{shoushtari2026snap} (training details in \cref{S-sec:family}, \cref{S-sec:impl}). The Lagrangian family differentiates in $t$ and contains LSD~\cite{boffi2025buildcm}. Shortcut~\cite{frans2025shortcut} and sCM~\cite{lu2025scm} are bridged by construction and have no island. A regression U-Net serves as a reference without spread. Finally, each region variant moves $5\%$ of the training pairs of the base sampler to one region of the triangle. The four regions are the \emph{corner point}, the \emph{launch strip}, the \emph{corner square} and the \emph{landing strip} of length $0.3$. Find details in \cref{S-sec:family} and \cref{S-sec:impl}.

\subsection{Overall performance}
\label{sec:exp-overall}

\paragraph{One step against 64 steps.}
Each network in \cref{tab:overall} is sampled with one step and with 64 steps, using the same weights. With 64 steps, every network is close to calibrated ($V/B=0.92$--$1.03$, gray columns), so the network itself has learned the uncertainty of the posterior. With no isthmus, the same networks in one step reach only $V/B=0.56$--$0.84$, so the uncertainty is lost in the one-step jump.

\paragraph{Quantitative comparison.}
\Cref{tab:overall} groups the methods by the time derivative in their objective, and its Bridge column marks how their training reaches the one-step corner. Without pairs near the corner ($\times$), the one-step samples lose spread and only partly follow their 64-step samples ($\rho\le0.70$), and $\rho$ varies most across training seeds. SNaP and the shortcut model (\cref{S-sec:impl}) place training pairs at or near the one-step pair by design (\brown). They follow their 64-step samples, but no method in this group is calibrated. With the isthmus (\bryes), every recipe reaches $\rho\ge0.93$, and MeanFlow and SNaP become the only one-step samplers that are both consistent with 64 steps and calibrated ($V/B=0.96$ and $0.95$). Their sample means are also more accurate than the output of the regression U-Net (single-sample metrics in \cref{S-sec:distortion}). LSD with the isthmus follows its 64-step samples but stays under-dispersed. The isthmus lowers PSNR$_\mu$ only for SNaP, by $0.06$\,dB, so calibration does not cost accuracy.

\paragraph{Qualitative comparison.}
In \cref{fig:qual}a, the one-step MRI sample of MeanFlow is noisier than its 64-step sample. Its per-pixel standard deviation over $16$ samples is scattered throughout the image, while that of 64 steps is concentrated on the edges of the anatomy. On CIFAR-10 (\cref{fig:qual}b), its one-step samples lose fine detail that the 64-step samples retain. With the isthmus, the one-step samples are close to the 64-step samples on both datasets, and on MRI their standard deviation maps are nearly identical.

\subsection{Placement of the supervised pairs}
\label{sec:exp-main}

\paragraph{Supervised region.}
\Cref{tab:mri} moves the same $5\%$ of training pairs between regions and keeps everything else fixed. The isthmus is the only region whose one-step samples are both consistent with 64 steps and calibrated, with $V_1/B_1=0.96$ on IXI as with 64 steps, and it is also the best region on the knee (\cref{S-tab:fastmri}). The other regions fail in two ways. Without extra pairs near the corner, the one-step samples match the 64-step samples in coarse structure but not in fine detail, where $\kappa_{\mathrm{f}}$ stays below $0.2$. With an island, the one-step samples follow the 64-step samples more closely than without extra pairs, but their spread shrinks, and $V_1/B_1$ falls to $0.26$ with the corner point and to $0.17$ with the launch strip. The landing strip, which covers the first $0.3$ of the landing edge, comes in second. The isthmus works best when replacing the island. With SNaP's corner square kept, $V/B$ reaches $0.80$, compared to $0.95$ when the isthmus replaces it (\cref{S-tab:snap-ixi}). Rescaling the one-step samples of an island to 64-step variance after training leaves $\rho$ unchanged and makes single samples perceptually worse (\cref{S-sec:mri}).

\begin{table}[t]
\centering
\footnotesize
\setlength{\tabcolsep}{2.4pt}\renewcommand{\arraystretch}{1.0}
\setlength{\aboverulesep}{0pt}\setlength{\belowrulesep}{0pt}%
\begin{tabular*}{\linewidth}{@{}c@{\hspace{3pt}}l@{\extracolsep{\fill}}ccccccc@{}}
\toprule
 & & \multicolumn{5}{c}{IXI brain (s.d.\ in \cref{S-tab:mri-full})\tst\bst} & \multicolumn{2}{c}{fastMRI knee} \\
\cmidrule(lr){3-7}\cmidrule(l){8-9}
\multicolumn{2}{@{}l}{Training pairs\tst\bst} & $\frac{B_1}{B_{64}}\!\downarrow$ & $\frac{V_1}{B_1}$ & $\rho$ & $\kappa_{\mathrm{c}}$ & $\kappa_{\mathrm{f}}$ & $\frac{V_1}{B_1}$ & $\kappa_{\mathrm{f}}$ \\
\midrule
\raisebox{-.35ex}{\includegraphics[height=2.15ex]{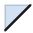}}\tst & no extra pairs, $p{=}1$ & 1.13 & 0.67 & 0.58 & \underline{0.94} & $-$0.03 & 0.72 & $-$0.31 \\
\raisebox{-.35ex}{\includegraphics[height=2.15ex]{figs/icons/ic_none.pdf}}\bst & no extra pairs, $p{=}0.5$ & 1.12 & 0.56 & 0.69 & 0.91 & 0.15 & 0.61 & $-$0.12 \\
\midrule
\raisebox{-.35ex}{\includegraphics[height=2.15ex]{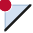}}\tst & corner point & 1.22 & 0.26 & 0.74 & 0.42 & 0.54 & 0.32 & 0.12 \\
\raisebox{-.35ex}{\includegraphics[height=2.15ex]{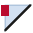}} & launch strip & 1.12 & 0.17 & 0.77 & 0.43 & 0.38 & 0.23 & 0.57 \\
\raisebox{-.35ex}{\includegraphics[height=2.15ex]{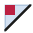}}\bst & corner square & \textbf{1.10} & 0.59 & \underline{0.90} & 0.92 & 0.53 & 0.59 & 0.53 \\
\midrule
\raisebox{-.35ex}{\includegraphics[height=2.15ex]{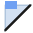}}\tst & landing strip & \textbf{1.10} & \underline{0.83} & \textbf{0.93} & \textbf{1.05} & \underline{0.79} & \underline{0.80} & \underline{0.66} \\
\rowcolor{ovours}\raisebox{-.35ex}{\includegraphics[height=2.15ex]{figs/icons/ic_edge.pdf}}\bst & \textbf{isthmus (ours)} & \underline{1.11} & \textbf{0.96} & \textbf{0.93} & 1.09 & \textbf{0.86} & \textbf{0.83} & \textbf{0.67} \\
\midrule
\tst\bst & \rf{64 steps} & \rf{1} & \rf{0.96} & \rf{1} & \rf{1} & \rf{1} & \rf{$\approx$1} & \rf{1} \\
\bottomrule
\end{tabular*}

\caption{\textbf{Effect of the supervised region on IXI and fastMRI.} Pairs along the landing edge keep the one-step samples consistent with 64 steps, and the full edge brings them closest to calibration.}
\label{tab:mri}
\end{table}

\paragraph{Supervised edge.}
\label{sec:exp-direction}
Which edge bridges the corner depends on the objective (\cref{prop:direction}). MeanFlow differentiates in the start time $r$, so its bridge is the landing edge. LSD~\cite{boffi2025buildcm} differentiates in the end time $t$, so its bridge is the launch edge. In \cref{tab:edgeswap} we move $5\%$ of pairs on either edge for both objectives (\cref{S-fig:direction}). Both edges add pairs next to the corner, and both raise $\rho$, but only the predicted edge brings the one-step variance close to the 64-step variance ($V_1/V_{64}=1.10$ for MeanFlow and $0.90$ for LSD). On the wrong edge, nothing in training sets the one-step variance (\cref{thm:island}), and it can end up too small or too large. MeanFlow stays as contracted as without extra pairs ($0.66$), while the one-step variance of LSD grows to $2.7$ times the 64-step variance.

\begin{table}[t]
\centering
\footnotesize
\setlength{\tabcolsep}{1.8pt}\renewcommand{\arraystretch}{1.08}
\setlength{\aboverulesep}{0pt}\setlength{\belowrulesep}{0pt}%
\begin{tabular}{@{}l@{\hspace{6pt}}c@{\hspace{4pt}}lcccc}
\toprule
Objective\tst\bst & \multicolumn{2}{l}{Edge pairs} & $B_1{\times}10^{3}\!\downarrow$ & $V_1/B_1$ & $V_1/V_{64}$ & $\rho$ \\
\midrule
MeanFlow\tst & \raisebox{-.35ex}{\includegraphics[height=2.15ex]{figs/icons/ic_none.pdf}} & none & 3.12{\tiny$\pm$0.02} & \underline{0.56}{\tiny$\pm$0.03} & \underline{0.66}{\tiny$\pm$0.05} & 0.69{\tiny$\pm$0.04} \\
(deriv.\ in $r$) & \raisebox{-.35ex}{\includegraphics[height=2.15ex]{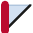}} & launch edge & \underline{3.11}{\tiny$\pm$0.05} & \underline{0.56}{\tiny$\pm$0.10} & \underline{0.66}{\tiny$\pm$0.12} & \underline{0.89}{\tiny$\pm$0.03} \\
\rowcolor{ovours}\bst & \raisebox{-.35ex}{\includegraphics[height=2.15ex]{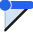}} & \textbf{landing edge} & \textbf{3.10}{\tiny$\pm$0.02} & \textbf{0.96}{\tiny$\pm$0.01} & \textbf{1.10}{\tiny$\pm$0.02} & \textbf{0.93}{\tiny$\pm$0.00} \\
\midrule
LSD\tst & \raisebox{-.35ex}{\includegraphics[height=2.15ex]{figs/icons/ic_none.pdf}} & none & \underline{3.19}{\tiny$\pm$0.05} & \underline{0.84}{\tiny$\pm$0.09} & \textbf{0.93}{\tiny$\pm$0.12} & 0.70{\tiny$\pm$0.04} \\
\rowcolor{ovours}(deriv.\ in $t$) & \raisebox{-.35ex}{\includegraphics[height=2.15ex]{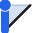}} & \textbf{launch edge} & \textbf{3.08}{\tiny$\pm$0.02} & \textbf{0.85}{\tiny$\pm$0.02} & \underline{0.90}{\tiny$\pm$0.03} & \textbf{0.96}{\tiny$\pm$0.00} \\
\bst & \raisebox{-.35ex}{\includegraphics[height=2.15ex]{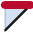}} & landing edge & 3.91{\tiny$\pm$0.15} & 1.92{\tiny$\pm$0.42} & 2.68{\tiny$\pm$0.52} & \underline{0.93}{\tiny$\pm$0.00} \\
\bottomrule
\end{tabular}

\caption{\textbf{Effect of the edge.} \textcolor{isth}{Blue} rows use edge \cref{prop:direction} predicts.}
\label{tab:edgeswap}
\end{table}

\begin{table}[t]
\centering
\footnotesize
\setlength{\tabcolsep}{2.1pt}\renewcommand{\arraystretch}{1.06}
\setlength{\aboverulesep}{0pt}\setlength{\belowrulesep}{0pt}%
\begin{tabularx}{\linewidth}{c>{\raggedright\arraybackslash}Xccccccc}
\toprule
 &  & \multicolumn{2}{c}{CelebA\tst\bst} & \multicolumn{5}{c}{CIFAR-10} \\
\cmidrule(lr){3-4}\cmidrule(l){5-9}
\multicolumn{2}{l}{Training pairs\tst\bst} & $\frac{B_1}{B_{64}}\!\downarrow$ & $\frac{V_1}{V_{64}}$ & FID$_1\!\downarrow$ & FID$_2\!\downarrow$ & $\rho$ & $\kappa_{\mathrm{f}}$ & Recall$\uparrow$ \\
\midrule
\raisebox{-.35ex}{\includegraphics[height=2.15ex]{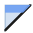}}\tst\bst & \rf{Dense ref.} & \rf{1.08} & \rf{1.10} & \rf{15.2} & \rf{10.9} & \rf{0.976} & \rf{0.87} & \rf{0.50} \\
\midrule
\raisebox{-.35ex}{\includegraphics[height=2.15ex]{figs/icons/ic_none.pdf}}\tst & MeanFlow & 1.40 & \underline{0.86} & 58.7 & 16.2 & 0.955 & 0.66 & 0.33 \\
\raisebox{-.35ex}{\includegraphics[height=2.15ex]{figs/icons/ic_point.pdf}} & + corner point & 1.42 & 0.37 & 25.7 & \underline{13.8} & 0.910 & 0.64 & \underline{0.41} \\
\raisebox{-.35ex}{\includegraphics[height=2.15ex]{figs/icons/ic_strip.pdf}} & + landing strip & \textbf{1.08} & 0.81 & \underline{20.0} & 14.1 & \textbf{0.965} & \underline{0.82} & \textbf{0.44} \\
\rowcolor{ovours}\raisebox{-.35ex}{\includegraphics[height=2.15ex]{figs/icons/ic_edge.pdf}}\bst & \textbf{+ isthmus (ours)} & \underline{1.11} & \textbf{0.91} & \textbf{19.2} & \textbf{12.5} & \underline{0.964} & \textbf{0.83} & \textbf{0.44} \\
\bottomrule
\end{tabularx}

\caption{\textbf{Inpainting on CelebA and unconditional generation on CIFAR-10.} On CelebA every entry uses time sampler of \cref{sec:exp-setup}, and on CIFAR-10 that of MeanFlow's ImageNet configuration, except Dense ref., a time sampler that stays dense up to landing edge.}
\label{tab:beyond}
\end{table}

\subsection{Analysis of islands and the isthmus}
\label{sec:exp-anatomy}

\paragraph{Start time of the jump.}
In \cref{fig:mech}a, a single jump starts at time $r$ on the 64-step trajectory and lands on the data. The isthmus keeps $\kappa$ close to $1$ for every start time. The other models fail only for jumps that start near the source. With the corner point, $\kappa$ is $0.42$ at $r=0$ and recovers above $0.9$ by $r=0.25$. In all four models of \cref{S-fig:heat}, the jump $(0,0.875)$, almost as long as the one-step jump, keeps $\kappa\ge0.84$, so failure is specific to the corner.

\paragraph{Number of sampling steps.}
In \cref{fig:mech}b, one step with the isthmus is as calibrated as 64 steps. The islands are miscalibrated with one step and close to calibrated with two, because the second jump starts at $r=0.5$, where the base sampler places many pairs near the landing edge.

\paragraph{Coarse and fine detail.}
In \cref{fig:mech}c, isthmus keeps $\rho\ge0.93$ in every frequency band. Without extra pairs near the corner, $\rho$ falls from $0.90$ in the coarsest to $0.21$ in the finest, so one-step samples lose fine detail first (spread in \cref{S-fig:mech-full}c).

\paragraph{Length of the supervised set.}
In \cref{fig:mech}d we grow the supervised set from the corner. Along the landing edge, $V_1/B_1$ rises with the length and reaches $0.95$ for the full edge, while the corner square stays below $0.7$ at every size. The share of pairs matters much less than the length. With $1\%$ of the pairs, $V_1/B_1$ is already $0.83$, and the results change little between $2.5\%$ and $10\%$ (\cref{S-fig:share}).

\paragraph{Longer training.}
\label{sec:exp-dyn}
An island learns from targets that no training pair ties to the data (\cref{sec:faces}), so its one-step sampler may change as training continues. We continue four MRI regions from $40$k to $120$k iterations (\cref{fig:mech}e, \cref{fig:teaser}d, \cref{S-tab:long}). The isthmus keeps $\rho$ between $0.92$ and $0.94$ throughout. With the corner point, the spread first shrinks further and then jumps to $1.40$ at $80$k, where $\rho$ falls to $0.59$. The landing strip, a partial bridge, holds only until $80$k. On CelebA, over a $4\times$ longer schedule, the one-step samples of the corner point turn into noise, with a spread $2.6$ times that of 64 steps, while its 64-step samples stay intact (\cref{fig:mech}f,g). The one-step quality of an island thus depends on when training stops, and agreement of isthmus does not (\cref{S-sec:dyn}).

\subsection{Generalization}
\label{sec:exp-beyond}

\begin{table}[t]
\centering
\footnotesize
\setlength{\tabcolsep}{1.9pt}\renewcommand{\arraystretch}{1.06}
\setlength{\aboverulesep}{0pt}\setlength{\belowrulesep}{0pt}%
\begin{tabularx}{\linewidth}{c>{\raggedright\arraybackslash}Xccccc}
\toprule
\multicolumn{2}{l}{Training pairs\tst\bst} & FID$_1\!\downarrow$ & FID$_2\!\downarrow$ & $\rho$ & $\kappa_{\mathrm{f}}$ & Recall$\uparrow$ \\
\midrule
\raisebox{-.35ex}{\includegraphics[height=2.15ex]{figs/icons/ic_none.pdf}}\tst & MeanFlow & 46.2{\tiny$\pm$12.6} & 32.8{\tiny$\pm$.4} & 0.961{\tiny$\pm$.006} & 0.84{\tiny$\pm$.09} & 0.30{\tiny$\pm$.04} \\
\rowcolor{ovours}\raisebox{-.35ex}{\includegraphics[height=2.15ex]{figs/icons/ic_edge.pdf}}\bst & \textbf{+ isthmus (ours)} & \textbf{34.8}{\tiny$\pm$1.5} & \textbf{26.2}{\tiny$\pm$1.2} & \textbf{0.963}{\tiny$\pm$.004} & \textbf{0.89}{\tiny$\pm$.04} & \textbf{0.37}{\tiny$\pm$.01} \\
\bottomrule
\end{tabularx}

\vspace{-.5em}
\caption{\textbf{Unconditional generation on ImageNet $32\times32$.} Best in \textbf{bold}. Results from 3 random seeds.}
\label{tab:i32}
\vspace{-.8em}
\end{table}

\paragraph{Inpainting on CelebA.}
On CelebA (\cref{tab:beyond}), the squared error of the one-step sample mean of MeanFlow is $40\%$ above that of 64 steps, and the isthmus lowers the gap to $11\%$. The corner point shrinks the one-step variance comparing to 64-step variance and leaves the error unchanged. With the isthmus, the one-step error and variance are both within about $10\%$ of the 64-step values, as with Dense ref., a time sampler that stays dense up to the landing edge (\cref{fig:mech}g).

\paragraph{Unconditional generation on CIFAR-10 and ImageNet.}
We start from the time sampler of MeanFlow's ImageNet configuration, which places almost no training pairs near the one-step corner (\cref{S-sec:impl}), and replace $5\%$ of its pairs with the isthmus. On CIFAR-10, the one-step FID~\cite{heusel2017fid} of the first seed drops from $58.7$ to $19.2$ (\cref{tab:beyond}, \cref{fig:qual}b). This approaches the $15.2$ of MeanFlow's CIFAR-10 sampler (Dense ref.), which already places many pairs near the corner. On ImageNet $32\times32$, the mean one-step FID drops from $46.2$ to $34.8$ (\cref{tab:i32}). The isthmus also makes the result far less dependent on the seed. On CIFAR-10 the one-step FID of two seeds differs by more than $20$ without the isthmus and by less than $1$ with it (\cref{S-tab:uncond-seeds}), and on ImageNet $32\times32$ its \textit{s.d.}\ over seeds falls from $12.6$ to $1.5$.

\section{Conclusion}
\label{sec:conclusion}
One-step flow maps trained by self-consistency can leave the pair they sample from unsupervised or on a supervision island, where the objective fixes neither the spread of the samples nor the course of training. The isthmus bridges this pair to the diagonal with a small change in the time sampler. 
It bounds the one-step error by the training residual, restores calibration, and ensures lasting agreement with the model’s own multi-step sampler. More broadly, the design of the time sampler is part of the objective: it decides which jumps training determines, and it can be checked before training.

\section*{Supplementary material}
Sections, lemmas, propositions, corollaries, remarks, tables and figures numbered
with the prefix S refer to the supplementary material, including the proofs, which will be
released upon acceptance.

{
    \small
    \bibliographystyle{ieeenat_fullname}
    \bibliography{refs}
}

\end{document}